\documentclass[pdflatex,sn-mathphys-num,oneside]{sn-jnl}% Basic Springer Nature Reference Style/Chemistry Reference Style
\usepackage{graphicx}
\usepackage[inkscapelatex=false]{svg}
\usepackage{subcaption}         % サブキャプション用
\usepackage{multirow}%
\usepackage{amsmath,amssymb,amsfonts}%
\usepackage{amsthm}%
\usepackage{mathrsfs}%
\usepackage[title]{appendix}%
\usepackage{xcolor}%
\usepackage{textcomp}%
\usepackage{manyfoot}%
\usepackage{booktabs}%
\usepackage{hhline}%
\usepackage{diagbox}%
\usepackage{algorithm}%
\usepackage{algorithmic}%
\usepackage{listings}%
\theoremstyle{thmstyleone}%
\theoremstyle{thmstyletwo}%

\theoremstyle{thmstylethree}%

\begin{document}

\title[Recycling computational processes of dynamic programming for combinatorial optimization problems: a reservoir computing approach]{Recycling computational processes of dynamic programming for combinatorial optimization problems: a reservoir computing approach}

\author*[1]{\fnm{Sora} \sur{Todaka}}\email{todaka.sora.62e@st.kyoto-u.ac.jp}

\author[1,2]{\fnm{Akihiro} \sur{Yamamoto}}\email{yamamoto.akihiro.5m@kyoto-u.ac.jp}

\author[1]{\fnm{Nozomi} \sur{Akashi}}\email{akashi.nozomi.2a@kyoto-u.ac.jp}

\affil*[1]{\orgdiv{Graduate School of Informatics}, \orgname{Kyoto University}, \orgaddress{\city{Sakyo-ku, Kyoto}, \country{Japan}}}

\affil[2]{\orgdiv{Institute for Liberal Arts and Sciences}, \orgname{Kyoto University}, \orgaddress{\city{Sakyo-ku, Kyoto}, \country{Japan}}}

\abstract{
Reusing previously computed results is a long-standing principle for reducing computational cost, but such reuse has largely been confined to a single problem's computation. 
Sharing computational processes across multiple simultaneously solved problems remains possible in principle, yet designing algorithms that exploit nontrivial cross-task relationships is difficult to do manually. Here, we use machine learning to discover such algorithms automatically.
Specifically, based on reservoir computing, we propose a method that uses computation results recorded by dynamic programming for combinatorial optimization problems as features for linear regression, leveraging them to assist other combinatorial optimization computations.
We validate the approach on the traveling salesman and subset sum problems.
Multiplexing the dynamic programming process improves approximation accuracy over generic features and reduces computation time compared with independent solutions.
These results suggest a new form of computation, distinct from conventional computational design, in which multiple processes efficiently share and recycle intermediate results and states.
}


\maketitle

\section{Introduction}
Saving computed results and reusing them efficiently has long been an important design principle for reducing computational cost.
Before the advent of computers, when computational resources were limited to human calculation, this principle was practiced rigorously.
For example, tables of trigonometric functions, logarithms, and probability density functions were indispensable for computing essential functions in daily life and engineering.
In particular, Napier's logarithm tables, completed after 20 years of computation, made it possible to reduce costly multiplication to table lookup and simple addition.

The idea of efficient reuse of computation persists in modern computer science in the form of lookup tables and memorization, and it underlies many approaches to computational efficiency.
For instance, dynamic programming (DP) records and reuses solutions to subproblems that appear repeatedly in inductive solution processes, and it is widely used in optimization problems as an algorithm design technique to eliminate redundant computation.
Another concrete example of the sharing intermediate results also appears in matrix multiplication.
While the naive algorithm for multiplying two $2\times2$ matrices requires eight multiplications, Strassen reduced this to seven by constructing seven shared intermediate terms for the entries.
More recently, AlphaTensor discovered new algorithms that reduce the number of multiplications for $4 \times 4$ matrix multiplication through reinforcement learning~\cite{fawzi}.
This is a striking example of machine learning autonomously uncovering complex sharing of intermediate results that would be difficult to design by hand.
It suggests that machine learning can enable more advanced and efficient sharing of computational resources.

Most of the efficiency gains discussed thus far concern reusing intermediate results within a single problem.
However, if we broaden the scope to computing multiple problems simultaneously, could we share their computation processes and further improve efficiency?
In fact, even for basic functions, simultaneous computation may be more efficient. A standard example is simultaneously finding the maximum and minimum of a sequence.
Computing these independently requires $N$ comparisons each, for a total of $2N$ comparisons.
Interestingly, computing them together can be done with only $1.5N$ comparisons~\cite{pohl1972sorting}.
This algorithm achieves this savings by performing $0.5N$ comparisons that contribute to both computations.
Such recycling of computational processes, which shares computation across multiple problems in this way, is considered to have a broader range of applications than the reuse of computation within a single problem.

In more complex settings, it is difficult to discover nontrivial relationships between problems and design how their computation processes can be shared.
There is a reservoir computing~\cite{jaeger2001echo, maass2002real, reservoircomputing} approach that exploit dynamics as a computational resource in a nontrivial way.
Reservoir computing uses a fixed dynamics called a ``reservoir'', driven by inputs that exploits the resulting complex dynamics as high-dimensional nonlinear features for computation.
A wide variety of dynamical systems can be used as computational resources, from physical phenomena~\cite{nakajimaPhysicalReservoirComputing2020a} such as liquid~\cite{liquid2002natschlager, fernando2003pattern, goto2021twin}, quantum~\cite{fujii2017harnessing, nakajima2019boosting, ghosh2019quantum, ghosh2019quantumb}, and robotic systems~\cite{hauser2011towards, caluwaerts2013locomotion,nakajima2013soft, zhao2013spine, nakajima2015information, akashi2024embedding} to discrete dynamical systems such as cellular automata~\cite{yilmaz2015symbolic,yilmaz2015machine, nichele2017deep, nichele2017reservoir, mcdonald2017reservoir, moran2019energy}.
These systems merely follow their inherent physical laws or rules, yet their evolution can be viewed as a nonlinear mapping of inputs into a high-dimensional feature space and leveraged for other computational purposes.
In other words, nontrivial relationships between the dynamics and the target computation are discovered and exploited through training of the output layer. Cellular automata also have a computational model aspect; one-dimensional cellular automata are known to be Turing complete~\cite{cookUniversalityElementaryCellular2004a}.
This suggests that the computation process itself can be multiplexed and repurposed for other objectives.

In this study, we explore the idea of recycling computational processes through reservoir computing in the context of combinatorial optimization algorithms.
The field of combinatorial optimization has been deeply studied alongside computational theory, and it provides guidance for analyzing shared computation processes.
As a new approach to reservoir computing, we propose a method that uses the DP process for solving one combinatorial optimization problem as a reservoir to solve another combinatorial optimization problem with the same input.
DP is a key algorithm design technique for optimization problems, and it explicitly stores intermediate values in a table for recycle. Therefore, we can define the computation process as the DP table and design the reservoir accordingly.
The proposed method computes a combinatorial optimization problem used for the reservoir (Problem A) and a target combinatorial optimization problem (Problem B) simultaneously: Problem A is solved by DP, and Problem B is solved by recycling the DP table for Problem A as features in linear regression.

We conducted numerical experiments using the traveling salesman problem (TSP) and subset sum problem (SSP), another representative combinatorial optimization problem, to evaluate the effectiveness of the approach.
As a result, we achieved higher accuracy in approximating those problems than with generic linear regression features and basic heuristics  by multiplexing the dynamic programming computation process, and we reduced computation time compared with solving the problems independently.

As Moore's law slows and the problems addressed in mathematical optimization continue to grow in scale, the finiteness of computational resources is again coming to the forefront.
In current computing practice, each time a problem is solved, a separate process is launched; once the computation ends, its process is discarded, and no sharing or recycle occurs.
Our approach suggests a different computational paradigm, in which multiple computational processes organically share and recycle intermediate results and states to make more efficient use of computational resources.

\section{Results}
\subsection{Proposed framework}
\begin{figure*}[t]
\centering
\includesvg[width=0.85\textwidth]{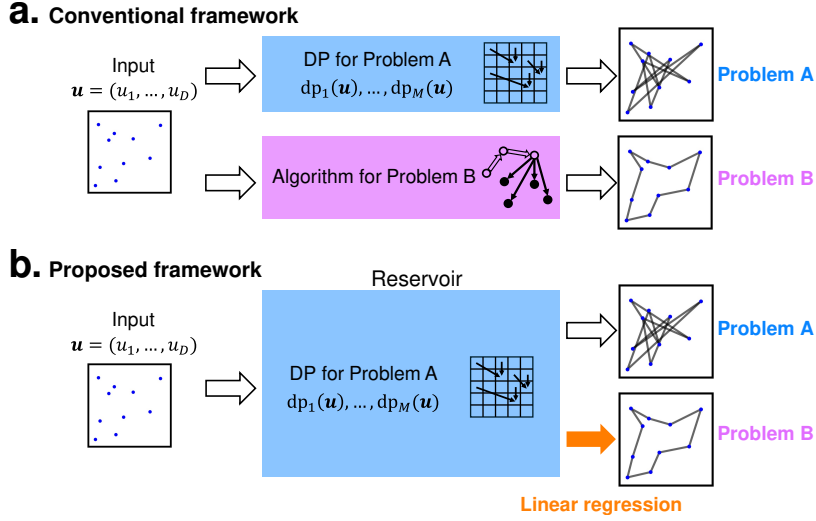}
\caption{\textbf{Conceptual comparison between the conventional and proposed frameworks.} \textbf{a}, In the conventional framework, Problem A and Problem B are each solved independently by a dedicated algorithm for the same input $\mathbf{u}$.
\textbf{b}, In the proposed framework, Problem A is solved by DP, and the resulting table $\mathrm{dp}_1(\mathbf{u}),\dots,\mathrm{dp}_M(\mathbf{u})$ is recycled as a reservoir: its entries serve as input features to a linear regression readout that approximates the solution of Problem B, eliminating the need to run a separate algorithm for Problem B.}
\label{fig:framework}
\end{figure*}

The method proposed in this study uses DP to solve a combinatorial optimization problem as a reservoir and restricts the trainable output layer to linear regression. 
Figure~\ref{fig:framework} shows conceptual comparison between the conventional and proposed frameworks.

\paragraph{Problem setting}
Let the input vector be $\mathbf{u}=(u_1, \dots, u_{D}) \in \mathbb{R}^D$, where $D$ is the dimension of the input.
We aim to solve multiple combinatorial optimization problems A, B, C, \dots \, for the same input simultaneously.
Conventionally, separate algorithms are designed for each problem and are solved independently.
In the proposed method, Problem A is solved by DP, and the other problems recycle the DP table, which records computation process for solving Problem A---as features to approximate the solutions by linear regression. This eliminates the need to run individual algorithms for problems B, C, \dots \, and replaces them with a lightweight linear transformation.

Suppose that running the DP for Problem A yields a table of size $M$. As each table element is a function of the input $\mathbf{u}$, we write
\begin{equation*}
    \mathbf{DP}(\mathbf{u}) = (\mathrm{dp}_1(\mathbf{u}), \dots, \mathrm{dp}_{M}(\mathbf{u}))^{\top} \in \mathbb{R}^M.
\end{equation*}

\paragraph{Linear readout}
In this method, $\mathbf{DP}(\mathbf{u})$ is shared as the feature vector for linear regression to solve another Problem B. Let the bias term be $b \in \mathbb{R}$, and define the weight vector as $\mathbf{W} = (w_1, \dots, w_M, b)^{\top} \in \mathbb{R}^{M+1}$. We also define the augmented feature vector by appending a constant 1,
$\boldsymbol{\phi}(\mathbf{u}) = (\mathrm{dp}_1(\mathbf{u}), \dots, \mathrm{dp}_M(\mathbf{u}), 1)^{\top} \in \mathbb{R}^{M+1}$. The model output $\hat{y}$ is computed by the following linear combination:
\begin{equation*}
    \hat{y} = \mathbf{W}^{\top}\boldsymbol{\phi}(\mathbf{u}) = \sum_{j=1}^{M} w_j \mathrm{dp}_j(\mathbf{u}) + b.
\end{equation*}
Here $\mathbf{DP}(\mathbf{u})$ is a function determined by the DP algorithm for solving Problem A and is fixed, rather than learned for Problem B. The only learnable parameters are the weights $\mathbf{W}$.

\paragraph{Training}
Training is performed in a supervised manner. Given training data consisting of $T_{\mathrm{train}}$ input-target pairs
$(\mathbf{u}^{(t)} \in \mathbb{R}^D, y^{(t)} \in \mathbb{R}) \quad (t=1, \dots, T_{\mathrm{train}})$,
we learn the weights $\mathbf{W}$ so that the model output $\hat{y}$ approximates the desired output $y$. The error is minimized using least squares with L2 regularization (ridge regression):
\begin{equation*}
    \mathbf{W} = \underset{\mathbf{W}}{\mathrm{argmin}} \left( \sum_{t=1}^{T_{\mathrm{train}}}(y^{(t)} - \hat{y}^{(t)})^2 + \lambda \|\mathbf{W}\|^2_2 \right).
\end{equation*}
Here $\lambda > 0$ is the regularization parameter.

\paragraph{Solution construction}
The proposed method is a model that approximates a real-valued function of the input. However, in optimization problems, in addition to estimating the optimal value of the objective function, one must construct decision variables that achieve that optimum (e.g., subsets or permutations in combinatorial optimization).
In this study, to construct solutions for combinatorial optimization problems using a real-valued function-approximation model, we adopt a DP-based framework. Specifically, we formulate the target problem as a DP and approximate the value function defined by the DP function with the learning model. Using the approximate values as guidance, we construct the solution sequentially.

In other words, solution construction can be reduced to multiple function-approximation problems. If the target problem can be reduced to $L$ functions $g_1(\mathbf{u}),\dots,g_L(\mathbf{u})$, then each $g_{\ell}$ is learned independently, and at the timing required for solution construction, its estimate is produced to construct the solution.

\subsection{Experimental setting}
The main purpose of this experiment is to examine how well a different problem can be solved by recycling, as features for linear regression, the DP computation process that solves a certain combinatorial optimization problem.
Followiing this objective, we conducted numerical experiments on the TSP and the SSP.
Both are fundamental problems in combinatorial optimization and have been studied extensively because of their broad applications\cite{matai2010traveling, hans2004knapsack}.
Moreover, since exact algorithms faster than brute-force search based on DP are known, these problems are suitable for the experimental setting in this study.

\subsubsection{Traveling salesman problem}
In the TSP, given a set of cities and travel costs between every pair of cities, the task is to find a tour that visits every city exactly once and returns to the starting city, with the minimum total travel cost.
The most naive algorithm enumerates all visit orders and requires $O(n!)$ time.
The TSP is NP-hard, and it is believed that no polynomial-time algorithm exists for finding an exact solution.
The well-known exact algorithm for the TSP is DP, which runs in $O(2^n n^2)$ time.
In addition, we also consider the problem of finding, among all tours that visit each city exactly once, the tour with the maximum travel cost.
This problem is also NP-hard and can also be solved by DP.
We call the problem of finding the maximum-cost tour MAXTSP, and the minimum-cost tour MINTSP.

\paragraph{Task setting}
The input is $n$ cities on the two-dimensional plane $[0,1)\times[0,1)$,
$\mathbf{u}=(x_1,y_1,\ldots,x_n,y_n)\in[0,1)^{2n}$,
and we use the Euclidean distance $d(i,j)=\sqrt{(x_i-x_j)^2+(y_i-y_j)^2}$ as the travel cost between two cities.
For this input, we aim to solve both MAXTSP and MINTSP simultaneously.
Specifically, we solve the DP for MAXTSP and use it as a computational resource to approximate MINTSP with the proposed method.
While, in principle, both would be solved independently by DP, in this study, we run DP only for MAXTSP (Problem A) and approximate MINTSP (Problem B) by linear regression to improve efficiency.

\paragraph{DP of MAXTSP as a reservoir}
Without loss of generality, we start the tour from city $1$.
Let $S\in 2^V$ be a subset of the city set $V=\{1,2,\ldots,n\}$ that contains city $1$, and let $v\in S$.
We define a state as the pair $(S,v)$.
Let $\mathrm{maxtsp}_{(S,v)}(\mathbf{u})\in\mathbb{R}$ be the maximum path length when starting from city $1$, having visited the set $S$ and currently being at city $v$.
The optimal value of MAXTSP is $\mathrm{maxtsp}_{(V,1)}(\mathbf{u})$.
The recurrence is
$\mathrm{maxtsp}_{(S,v)}(\mathbf{u})=\max_{w\in S,\ w\neq v} d(w,v) + \mathrm{maxtsp}_{(S\setminus\{v\},w)}(\mathbf{u})$.
We use the $O(2^n n)$ dimensional table $\mathrm{maxtsp}_{(S,v)}(\mathbf{u})$ as features for linear regression.

\paragraph{Data generation}
The input is generated by sampling each city coordinate $(x_i,y_i)\in [0,1) \times [0,1)$ independently from the uniform distribution.
Since the input is originally a set of $n$ points, we sort the points in lexicographic order $\prec$ (first by $x$, and then by $y$ if $x$ is equal) to make the representation unique.
Thus, $(x_1,y_1)\prec(x_2,y_2)\prec\cdots\prec(x_n,y_n)$ holds.
Among the generated instances, $T_{\mathrm{train}}$ are used for training and $T_{\mathrm{eval}}$ for evaluation.

In the experiments, we perform two tasks: (i) prediction of the optimal value of MINTSP and (ii) construction of a tour.

\paragraph{(i) Prediction of the optimal value}
The objective function is the optimal value of MINTSP.
Evaluation is performed by the mean absolute percentage error (MAPE) between the predicted and true values on the evaluation data.
We set $n=10,\lambda=0.01,\; T_{\mathrm{train}}=10^5,$ and  $T_{\mathrm{eval}}=10^4$.
In this experiment, we use the DP table for MAXTSP as features for linear regression to predict the optimal value of MINTSP.
To compare the feature performance, we also evaluate linear regression with generic features as baselines: linear regression on the raw input (INPUT), extreme learning machine (ELM)~\cite{huang2004extreme}, and next generation reservoir computing (NG-RC)~\cite{ngrc}, the last of which we examine in two variants, NG-RC-INPUT and NG-RC-DIST.
In addition, we compare the model with three further methods that use features specifically related to MINTSP: $\sqrt{NA}$~\cite{beardwood1959shortest}, $\mathrm{std_{RT}}$~\cite{kouOptimalTSPTour2022a}, and minimum-weight matching (MATCHING).
Details of all comparison methods are given in Methods.

\paragraph{(ii) Solution construction}
Beyond predicting the optimal value, we also construct a tour (solution).
We evaluate the average percentage by which the tour length produced by the model exceeds the optimal solution over all evaluation data.
We set $n=14,\lambda=0.01,\; T_{\mathrm{train}}=5\times10^5$, and $T_{\mathrm{eval}}=10^4$.

To reduce the computation time in this experiment, we performed feature selection during training and used only 2,000 features instead of the full MAXTSP table with size 53,500 features.
For feature selection, we used recursive feature elimination~\cite{guyon2002gene} and reduced the number of features by $10\%$ at each step.
We provide detail analysis of feature selection in Appendix.

In this experiment, the model outputs an approximate solution given an input.
To assess solution quality, we compare the model with three representative TSP approximation algorithms: nearest neighbor (NN), Christofides algorithm (CHRIS)~\cite{karlinSlightlyImprovedApproximation2021}, and farthest insertion (INSERT); details are given in Methods.

\paragraph{Out-of-distribution evaluation (TSPLIB)}
To examine generalization performance outside the training data distribution, we also performed evaluations on instances from TSPLIB~\cite{reineltTSPLIBTravelingSalesman1991}, which is a standard benchmark for TSP.
This dataset is constructed from real cities.
We performed the evaluation on burma14, the smallest TSPLIB instance with 14 cities.

\subsubsection{Subset sum problem}
The SSP involves, given a set of numbers $S$ and a target value $n$, finding a subset of $S$ whose sum equals $n$.
In this study, we distinguish the following three tasks:
\begin{itemize}
    \item \textbf{DECSSP$_n$}: Decide whether there \textbf{exists} a subset whose sum equals $n$.
    \item \textbf{MAXSSP$_n$}: If such a subset exists, find the \textbf{maximum} number of elements among those subsets.
    \item \textbf{MINSSP$_n$}: If such a subset exists, find the \textbf{minimum} number of elements among those subsets.
\end{itemize}
DECSSP is easier than MAXSSP and MINSSP because it requires only an existence check.
For example, for $S=(2,5,8,4,2)$ and $n=13$, a subset summing to $13$ exists; the one with the maximum number of elements is $(2,5,4,2)$, and the one with the minimum number of elements is $(5,8)$.
Therefore MAXSSP$_{13}(S)$ is $4$ and MINSSP$_{13}(S)$ is $2$.
On the other hand, when $n=18$, no such subset exists.
SSP is NP-hard, and a pseudo-polynomial-time DP algorithm is known to be an exact method that is faster than brute-force search.

\paragraph{Task setting}
The input is $\mathbf{u}=(u_1,\ldots,u_{D})\in\{1,\ldots,U\}^{D}$, where $U$ is the maximum input value and $D$ is the input dimension.
For this input, we aim to solve MAXSSP and MINSSP simultaneously.
Rather than solving them independently, we first solve the simpler common subproblem DECSSP by DP and use its table as a computational resource to solve MINSSP and MAXSSP with the proposed method.
Compared with computing each independently, this approach shares the computation process of the common DECSSP subproblem and improves efficiency.

\paragraph{DP of DECSSP as a reservoir}
Let $\mathrm{decssp}_{i,j}(\mathbf{u})$ denote whether there exists a subset of $(u_1,u_2,\cdots,u_{i})$ whose sum is $j$.
We fill the table in increasing order of $i$ and $j$ for $1\leq i\leq D$ and $0\leq j\leq UD$ using the following recurrence:
$\mathrm{decssp}_{i,j}(\mathbf{u}) = \mathrm{decssp}_{i-1,j}(\mathbf{u}) \lor \mathrm{decssp}_{i-1,j-u_i}(\mathbf{u})$.
This recurrence considers two cases depending on whether the $i$-th element is used:
if we exclude $u_i$, we reduce to $\mathrm{decssp}_{i-1,j}(\mathbf{u})$; if we include it, we reduce to $\mathrm{decssp}_{i-1,j-u_i}(\mathbf{u})$.
We use this $UD^2$-sized table $\mathrm{decssp}_{i,j}(\mathbf{u})$ as features for linear regression.

\paragraph{Data generation}
We generate $D$ integers independently and uniformly at random between $1$ and $U$, and sort the values in descending order to form one instance.
Thus $u_1 \geq u_2 \geq \cdots \geq u_D$ holds.
Among the generated instances, $T_{\mathrm{train}}$ are used for training and $T_{\mathrm{eval}}$ for evaluation.

We again conduct two experiments: (i) prediction of the optimal value and (ii) solution construction.

\paragraph{(i) Optimal value prediction}
The objective is the number of elements of MAXSSP$_n$ and MINSSP$_n$ when the sum $n$ is specified between $1$ and $UD$.
For evaluation, we round the model output to an integer and compute the accuracy---that is, whether it matches the integer answer.
We report Cohen's Kappa normalized to $[0,1]$, where $0$ corresponds to the most naive model that always outputs the single value with the highest frequency in the training data.
Here $p_e$ is the accuracy of that naive model.
$\mathrm{Kappa}:=1-\frac{1-\mathrm{accuracy}}{1-p_e}$.
In this experiment, when predicting MAXSSP$_n$ and MINSSP$_n$, we first predict whether a subset summing to $n$ exists; if we predict it exists, we then predict the number of elements.
We set $U=10,D=10,\lambda=0.01$, and $T_{\mathrm{eval}}=2,000$, with $T_{\mathrm{train}}$ set to $10$ times the number of features for each method, and vary the remaining three parameters.
To compare the DP table of DECSSP as features, we also use the generic features INPUT, ELM, and NG-RC as baselines, as in the TSP experiments.

\paragraph{(ii) Solution construction}
We also construct solutions rather than merely predicting optimal values.
Evaluation is the accuracy of correctly constructing a subset with the maximum (minimum) number of elements.
We perform experiments for all $n$ between $1$ and $UD$.
We set the same parameters as (i).
In this experiment, we compare the computation time with the case where MAXSSP and MINSSP are solved independently by DP.
In the proposed method, we first construct the DP table for solving DECSSP and then use this DP table as features to construct solutions for MAXSSP and MINSSP.

\subsection{Experimental results}
\subsubsection{Traveling salesman problem}
\begin{figure}[ht]
\centering
\includesvg[width=1.0\linewidth]{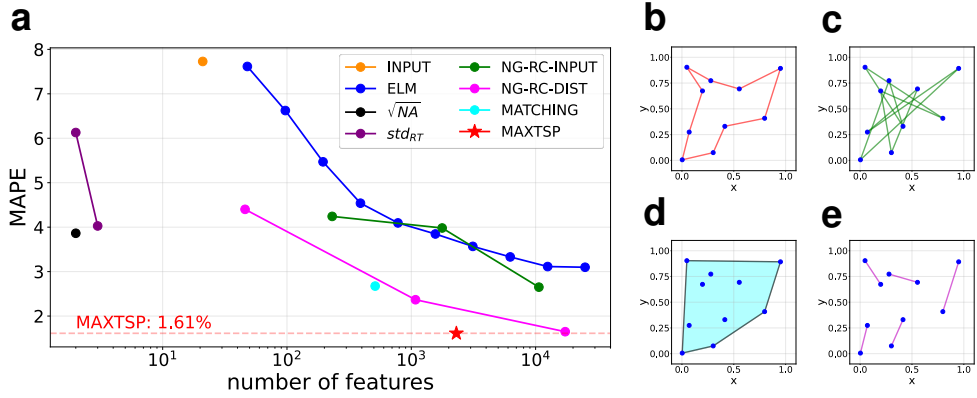}
\caption{\textbf{The performances of MINTSP optimal value prediction and examples of solutions.} \textbf{a}, the mean absolute percentage error (MAPE) of the predicted MINTSP optimal value as a function of the number of features, for the proposed method (MAXTSP) and baseline features (INPUT, ELM, $\sqrt{NA}$, $\mathrm{std_{RT}}$, NG-RC-INPUT, NG-RC-DIST, MATCHING). \textbf{b}--\textbf{e}, Examples, for the same instance, of the solutions computed by each method for use in optimal value prediction. \textbf{b}, The target MINTSP tour. \textbf{c}, The MAXTSP tour computed by the DP used as the reservoir in the proposed method. \textbf{d}, The convex hull used to compute the $\sqrt{NA}$ feature. \textbf{e}, The minimum-weight matching computed by DP for the MATCHING baseline.}
\label{fig:tsp_value}
\end{figure}
\paragraph{(i) Optimal value prediction}
The results of the optimal value prediction experiment are shown in Fig.~\ref{fig:tsp_value}.
A larger number of features (i.e., further to the right in the plot) corresponds to higher expressive power and better performance. 
Indeed, for ELM and NG-RC, increasing the number of features reduces the error.
The proposed method using the MAXTSP DP table as features achieves an average error of 1.61\%, which is the highest performance in other methods.
In particular, reaching similar performance with NG-RC-DIST, which we consider the most suitable baseline for approximating TSP, requires approximately 10 times more features, indicating that the MAXTSP table contains many features that are useful for approximating MINTSP.
Looking at MATCHING, where a different algorithm is used as the reservoir, the error is lower than NG-RC-INPUT and ELM at a comparable feature count, but it is still worse than MAXTSP.
This suggests that for approximating MINTSP, the DP for MAXTSP carries information processing more closely aligned than the DP for MATCHING.

\begin{table}[htbp]
\centering
\caption{\textbf{Results of constructing approximate MINTSP tours.} For each method, the average percentage by which the constructed tour is longer than the optimal solution (Gap) and the average time required to construct the solution (Time) are reported per test instance.}
\label{tab:tsp_tour_construction}
\begin{tabular*}{0.8\linewidth}{@{\extracolsep{\fill}}lcc}
\toprule
Method & Gap(\%) & Time (ms) \\
\midrule
NN     & 14.3  & 0.004   \\
CHRIS  & 18.8  & 0.042   \\
INSERT & 1.28  & 0.020  \\
\textbf{MAXTSP} & 1.03  & 1.38   \\
\midrule
DP (exact)     & 0.00  & 11.9   \\
\bottomrule
\end{tabular*}
\end{table}

\paragraph{(ii) Solution construction}
The results of tour construction are shown in Table~\ref{tab:tsp_tour_construction}.
For the evaluation data, we report the average percentage by which each method's constructed tour is longer than the optimal solution and the average computation time per instance.
The proposed MAXTSP method achieved the best performance among the compared approximation algorithms, with an error of 1.03\%.
It was also about nine times faster than independently computing MINTSP by DP.
An important point is that the compared algorithms are designed to solve MINTSP, whereas in the proposed method, the features come from a DP designed for MAXTSP, and only the linear regression component is designed for MINTSP.
This indicates that the DP designed for MAXTSP contains enough information processing capacity to yield approximations that outperform heuristic algorithms dedicated to MINTSP.
\begin{figure}[htbp]
\centering
\includesvg[width=1.0\linewidth]{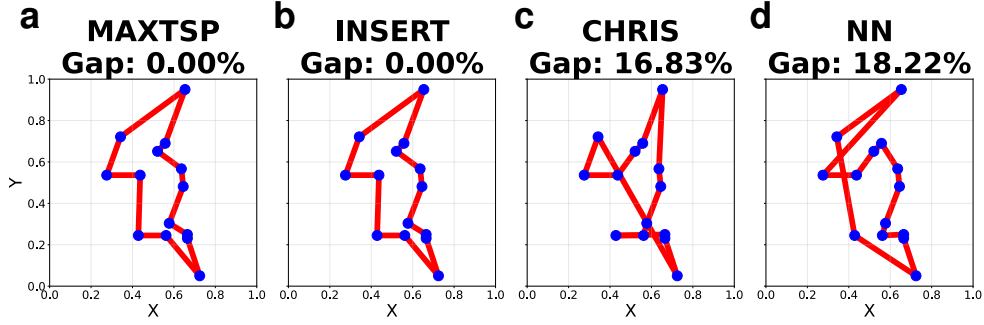}
\caption{\textbf{Results for the burma14 instance.} Tours constructed for the TSPLIB instance burma14 by the proposed method (\textbf{a}, MAXTSP) and the baseline heuristics (\textbf{b}, INSERT; \textbf{c}, CHRIS; \textbf{d}, NN), with the percentage gap from the optimal tour length shown above each panel.}
\label{fig:burma14}
\end{figure}
Fig~\ref{fig:burma14} shows the actual tours constructed by each method for the TSPLIB instance burma14.
The proposed MAXTSP achieved zero error and constructed the optimal solution.
This indicates that the model trained on synthetic data generalizes well to out-of-distribution real data.

\subsubsection{Subset sum problem}
\begin{figure}[htbp]
\centering
\includesvg[width=1.0\linewidth]{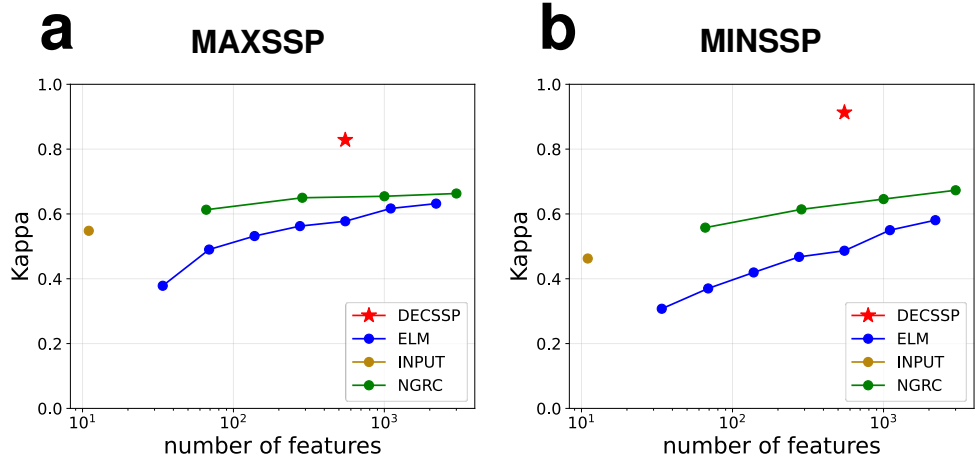}
\caption{\textbf{The performances of SSP optimal value prediction.} \textbf{a}, \textbf{b}, Kappa of the predicted optimal value for \textbf{a}, MAXSSP and \textbf{b}, MINSSP on the evaluation data, as a function of the number of features, for the proposed method using the DECSSP DP table (DECSSP) and baseline features (INPUT, ELM, NG-RC).}
\label{fig:ssp_value}
\end{figure}

\newlength{\mycolwidth}
\settowidth{\mycolwidth}{Accuracy(\%)}
\begin{table*}[htbp]
\centering
\caption{\textbf{Results of SSP solution construction.} For each component of the conventional approach (independent DP for MAXSSP and MINSSP) and the proposed approach (a shared DP for DECSSP followed by readout layers for MAXSSP and MINSSP), the accuracy of the constructed solution and the average time required per instance are reported; DP for DECSSP alone does not yield a solution and therefore has no associated accuracy. The rightmost Accuracy and Time columns report, respectively, the accuracy averaged over the solved tasks and the total time summed over all components, giving the overall cost of obtaining both the MAXSSP and MINSSP solutions.}
\label{tab:ssp_construction}
\begin{tabular}{|c|l|c|c||c|c|}
\hline
 & Component & Accuracy& Time & Accuracy & Time \\
\hline
\multirow{2}{*}{\begin{tabular}{c}Conventional\end{tabular}}
 & DP for MAXSSP & 100 & 0.295 & \multirow{2}{*}{100} & \multirow{2}{*}{0.579} \\
\hhline{~---~~}
& DP for MINSSP & 100 & 0.284 &  &  \\
\noalign{\hrule height 1.2pt}
\multirow{3}{*}{\begin{tabular}{c}\textbf{Proposed}\end{tabular}}
 & DP for DECSSP & \diagbox[height=\line,width=\mycolwidth]{}{}& 0.026 & \multirow{3}{*}{95.7} & \multirow{3}{*}{0.065} \\
\hhline{~---~~}
 & Readout for MAXSSP & 91.8 & 0.019 &  &  \\
\hhline{~---~~}
 & Readout for MINSSP & 99.5 & 0.021 &  &  \\
\hline
\end{tabular}
\end{table*}
\paragraph{(i) Optimal value prediction}
The results of the optimal value prediction experiment are shown in Fig.~\ref{fig:ssp_value}.
While more features generally imply higher expressive power and better performance, ELM and NG-RC indeed improved in accuracy as the number of features increases.
The proposed method using the DECSSP DP table as features achieved the best Kappa for both the MAXSSP and MINSSP tasks, showing excellent performance per feature count.
For both tasks, it outperformed generic-feature baselines even when those used more than 10 times as many features.
This indicates that the DECSSP DP table is a good feature set for both MAXSSP and MINSSP.
We provide the results of how performance changed when the parameter U and D were varied in Appendix.

\paragraph{(ii) Solution construction}
The results of solution construction are shown in Table~2.
The proposed method correctly constructed solutions with an accuracy of 91.8\% for MAXSSP and an accuracy of 99.5\% for MINSSP.
In the conventional approach, MAXSSP and MINSSP are solved independently by DP.
In the proposed method, we compute the lighter DP for DECSSP and share its table as linear regression features for MAXSSP and MINSSP.
Thus the computations required to solve the two tasks are the DP for DECSSP and the output layers for MAXSSP and MINSSP.
Since each of these three computations is lightweight, we can obtain approximate solutions faster than the conventional exact DP approach.
We provide the detail results of the solution construction in Appendix.

\section{Discussion}
Inspired by reservoir computing, this study proposed a method that treats algorithms as more general computational resources and recycles computation processes in combinatorial optimization problems in a multiplexed manner.
To verify its effectiveness, we conducted experiments on the TSP and the SSP.
For the TSP, we showed that the DP process for one problem (MAXTSP) provides features that are useful for solving a different problem (MINTSP), even though the two objectives are, superficially, opposites of each other.
For the SSP, we showed that sharing the computation process of a simpler common subproblem (DECSSP) makes it possible to solve two related problems (MAXSSP and MINSSP) at a substantially lower additional cost than solving each of them independently.
Together, these results indicate that the DP table computed for one problem is not merely a byproduct of solving that problem but a reusable computational resource that can be repurposed, through a lightweight linear readout, to solve other problems that share the same input.

Within the same framework, the proposed method can be extended in two directions.
First, the reservoir is not restricted to DP: any algorithm that exposes an intermediate computation process, such as a table, a set of internal variables, or a sequence of states, can in principle serve as the reservoir.
Second, the readout is not restricted to linear regression: replacing it with a more expressive output model, such as a nonlinear or hierarchical readout, may make it possible to recycle computational resources whose relation to the target problem is more complex than what a linear map can capture.

\paragraph{Relation to machine learning for combinatorial optimization}
Solving combinatorial optimization problems with machine learning has already been studied extensively~\cite{bengio2021machine}.
Such approaches offer several advantages beyond simply improving performance: a single trained model can be applied to a variety of problems, and learning can automatically capture the distribution of instances and problem-specific characteristics that would be difficult to build into a hand-designed algorithm.
However, this line of work is, almost without exception, aimed at improving the performance of the solution algorithm for an individual problem considered in isolation.
The present study takes a different stance: rather than designing an independent solution algorithm for each problem, we aim to share computational processes across multiple problems solved for a common input.
In other words, our focus is on how much additional information can be extracted from an existing algorithm's computation, and at how little additional cost, rather than on how much a single algorithm's own performance can be improved.

For this reason, our approach is complementary to, rather than competing with, research on improving individual algorithms: whenever a new, higher-performing algorithm for some problem is developed, our framework offers a way to ask how much additional information about \emph{other} problems can be extracted from that algorithm's computation process.
This is also why, throughout our experiments, we compared the proposed method against baselines that solve each problem independently or that use generic features, rather than against the best-known specialized algorithm for each target problem; our aim was to evaluate the value of recycled computation, not to outperform the state of the art on any single problem.

\paragraph{Engineering and scientific implications}
On completion, this line of research is expected to yield, from an engineering standpoint, a technique for obtaining information about several related problems from a single optimization computation at low additional cost, providing a basis for accelerating computation through the sharing of computational processes.
Of course, the proposed method does not work well for an arbitrary combination of algorithm and problem: because the readout is kept lightweight, most of the computation needed to solve the target problem must be carried by the computational process of the original algorithm, so the algorithm and the target problem must be substantively related. This raises the question of which combinations of algorithm and problem allow the method to succeed.

Addressing this question, from a scientific standpoint, this line of research is expected to lay the foundation for a new framework that treats the relationship between a problem and an algorithm as an object of study in its own right, potentially leading to the discovery of nontrivial relationships between problems that are not apparent from their definitions alone, such as the relationship we observed between MAXTSP and MINTSP.

More broadly, the pursuit of extensive sharing of computational processes suggests a computational paradigm distinct from current practice, in which a separate process is launched each time a problem is solved and is discarded once the computation ends, with no sharing or recycle across processes.
Instead, our results point toward a mode of computation in which multiple computational processes organically share and recycle intermediate results and states, analogous to how the biological brain flexibly repurposes shared neural resources across different cognitive tasks, thereby making more efficient use of computational resources.

\newcommand{\argmax}{\mathop{\rm arg~max}\limits}
\newcommand{\argmin}{\mathop{\rm arg~min}\limits}
\section*{Methods}
The execution environment consisted of Intel Xeon Gold 6230 CPU for TSP and Intel Xeon w3-2535 CPU with an NVIDIA RTX 2000 Ada Generation GPU for SSP.

\subsection*{Comparison methods}
Here we describe in detail the baseline and comparison methods used in the experiments.

\paragraph{Generic regression features}
For the optimal value prediction experiment, we compare the proposed DP features against the following generic features.
\begin{itemize}
    \item \textbf{Linear regression on the raw input (INPUT):}
    Use $\mathbf{u}=(u_1,\ldots,u_{D})$ as features.
    The number of features is $D$.

    \item \textbf{Extreme learning machine (ELM)}~\cite{huang2004extreme}\textbf{:}
    A single-hidden-layer feedforward neural network where only the output layer is trained and the other parameters are randomly initialized.
    It can be regarded as linear regression on random nonlinear features.
    The number of features $M$ can be set freely, and, in general, a larger $M$ yields higher expressive power~\cite{huang2006extreme}.

    \item \textbf{Next generation reservoir computing (NG-RC)}~\cite{ngrc}\textbf{:}
    Multivariate polynomial regression that uses all terms up to the degree $p$ of the input variables as features.
    We examine two variants: polynomials of the input variables (NG-RC-INPUT) and polynomials of the distance matrix between cities (NG-RC-DIST).
    The number of features is $\binom{D+p}{p}$ where $D$ is the number of variables, and the maximum degree $p$ can be chosen freely.
    Larger $p$ yields higher expressive power.
\end{itemize}

\paragraph{TSP-specific comparison features}
In addition, we compare the model with the following three methods as features related to MINTSP.
\begin{itemize}
    \item {$\mathbf{\sqrt{NA}}$:}
    A classical and well-known method to approximate the optimal value of MINTSP by linear regression.
    Let $N$ be the number of vertices and $A$ the area of the convex hull of the point set, and use $\sqrt{NA}$ as the feature~\cite{beardwood1959shortest}.
    The number of features is one.

    \item {$\mathbf{std_{RT}}$:}
    To predicting the MINTSP optimal value in settings not limited to the Euclidean plane, a method was proposed that uses the standard deviation of random tour lengths as a feature~\cite{kouOptimalTSPTour2022a}.
    Theoretically, this value is linearly dependent on $\sqrt{NA}$ under certain conditions.
    In this experiment, we use two features: the standard deviation $\mathrm{std_{RT}}$ and the mean $\mathrm{mean_{RT}}$ of random tour lengths.

    \item \textbf{Minimum-weight matching (MATCHING):}
    When $n$ is even, consider forming $n/2$ pairs of points without using any vertex twice, minimizing the sum of distances within each pair.
    We solve this problem by DP and use its table as features.
    This is a variant of the proposed method, intended to examine the case where the DP solves a problem other than MAXTSP.
\end{itemize}

\paragraph{TSP approximation algorithms}
For the solution construction experiment, we compare the model with the following representative TSP approximation algorithms.
\begin{itemize}
    \item \textbf{Nearest neighbor (NN):}
    A greedy method that repeatedly visits the nearest vertex.
    The time complexity is $\mathrm{O}(N^2)$.

    \item \textbf{Christofides algorithm (CHRIS):}
    An algorithm with a 1.5-approximation guarantee when distances satisfy the triangle inequality.
    It had the best known guarantee until 2021~\cite{karlinSlightlyImprovedApproximation2021}.
    The time complexity is $\mathrm{O}(N^3)$.

    \item \textbf{Farthest insertion (INSERT):}
    An algorithm that starts with the tour of the two farthest vertices and inserts the farthest remaining vertex into the best position in the tour.
    The time complexity is $\mathrm{O}(N^2)$.
\end{itemize}

\subsection*{Training details of SSP}
In the SSP experiments, the objective function MAXSSP$_n$ takes values in $\{1, \ldots, D, \mathrm{false}\}$.
If no subset with sum $n$ can be formed, the value is $\mathrm{false}$; otherwise, it is the maximum subset size in $\{1, \ldots, D\}$.
To learn these two types of values, $\mathrm{false}$ and size, by linear regression, we train two types of weights and predict the value of MAXSSP$_n$.
The first weight learns whether the output is $\mathrm{false}$, and the second weight learns the size using only training data whose labels are not $\mathrm{false}$.
At prediction time, we first use the first weight to predict whether the output is $\mathrm{false}$.
If it is predicted not to be $\mathrm{false}$, we then use the second weight to predict the size.
The prediction of MINSSP$_n$ is performed in the same way.

\subsection*{Solution construction}
We describe in detail what the model learns in each experiment and how it constructs a solution.

\paragraph{Solution construction for MINTSP}
The objective function learned by the model is $g_{(n,v)}(\mathbf{u})$, which is the minimum length of a path that starts from vertex $1$ in an $n$-vertex input $\mathbf{u}$, visits every vertex exactly once, and ends at vertex $v$.
To approximate this function, the model uses the DP table for solving MAXTSP as features.
Let $\mathcal{I}=\{(S,v) \mid S\in 2^V,\ 1\in S,\ v\in S\}$ be the set of valid states; that is, $S$ ranges over the $2^{n-1}$ subsets of $V$ that contain city $1$, and for each such $S$, $v$ ranges over its $|S|$ elements.
Fixing an arbitrary enumeration $(S_1,v_1),(S_2,v_2),\ldots,(S_{|\mathcal{I}|},v_{|\mathcal{I}|})$ of $\mathcal{I}$, we define the vector
$\mathbf{MAXTSP}(\mathbf{u}):=\big(\mathrm{maxtsp}_{(S_1,v_1)}(\mathbf{u}),\mathrm{maxtsp}_{(S_2,v_2)}(\mathbf{u}),\ldots,\mathrm{maxtsp}_{(S_{|\mathcal{I}|},v_{|\mathcal{I}|})}(\mathbf{u})\big)\in\mathbb{R}^{|\mathcal{I}|}$.
Using this vector as features, the model learns the weight $\mathbf{W}_{(n,v)}$ and approximates
$\mathbf{W}_{(n,v)}\mathbf{MAXTSP}(\mathbf{u}) \approx g_{(n,v)}(\mathbf{u})$.

The algorithm for constructing a solution is greedy.
Starting from vertex $1$, it greedily determines the next vertex to visit in sequence.
Suppose that the set of already visited vertices is $S \in 2^V$ and the current vertex is $v$.
Let $T=V\setminus S$ and $k=|T|$.
For a subset $R=\{i_1<i_2<\cdots<i_m\}\subseteq V$, with indices listed in increasing order, we also write
$\mathbf{u}[R]=(x_{i_1},y_{i_1},x_{i_2},y_{i_2},\ldots,x_{i_m},y_{i_m})$.
Since $1$ is the smallest element of $V$, whenever $1\in R$ it is always mapped to the first coordinate pair, so $g_{(m,\cdot)}(\mathbf{u}[R])$ is consistent with the convention that the path starts from vertex $1$.
The next vertex to visit is determined by
$\argmin_{w \in T} d(v,w)+g_{(k+1,w)}(\mathbf{u}[T \cup \{1\}])$.
This represents the length of the remaining path when vertex $w$ is visited next in the tour.
In practice, $g_{(k+1,w)}(\mathbf{u}[T \cup \{1\}])$ is replaced by the value predicted by linear regression using $\mathbf{MAXTSP}(\mathbf{u})$ as features.
Specifically, we use $\mathbf{W}_{(k+1,w)}\mathbf{MAXTSP}(\mathbf{u}[T \cup \{1\}])$ in place of $g_{(k+1,w)}(\mathbf{u}[T \cup \{1\}])$.
If these predictions are correct, the optimal solution is obtained.
The table $\mathbf{MAXTSP}(\mathbf{u}[T \cup \{1\}])$ is a subset of $\mathbf{MAXTSP}(\mathbf{u})$, and the corresponding portion is extracted from $\mathbf{MAXTSP}(\mathbf{u})$ and used as features.

\paragraph{Solution construction for SSP}
In the solution construction experiment for MAXSSP, the model learns the objective function
$\mathrm{maxssp}_{i,j}(\mathbf{u})=\mathrm{MAXSSP}_{j}(\mathbf{u}[\{1,\ldots,i\}])\in \{1,\ldots,i,\mathrm{false}\}$
for $1 \leq i \leq D$ and $0 \leq j \leq UD$.
As described in ``Training details of SSP'' above, two types of weights are learned and used during training and prediction.
The solution is constructed by the following greedy method, which decides in order from the $D$-th element whether each element is included in the subset.
Suppose that we have decided whether to include elements up to the $(i+1)$-th element and that the sum of the selected subset is $j$.
Then, whether to include the $i$-th element is determined by comparing $\mathrm{maxssp}_{i-1,n-j}(\mathbf{u})$ and $\mathrm{maxssp}_{i-1,n-j-u_i}(\mathbf{u})+1$.
If the former is larger, the element is not included in the subset; if the latter is larger, the element is included.
MINSSP is handled in the same way.

\bmhead{Acknowledgements}
This work supported by JSPS KAKENHI Grant No. 25K00011.
\newpage
\appendix
\def\thesection{Appendix \Alph{section}}
\section{Training size}
\begin{figure}[htbp]
\centering
\includesvg[width=1.0\linewidth]{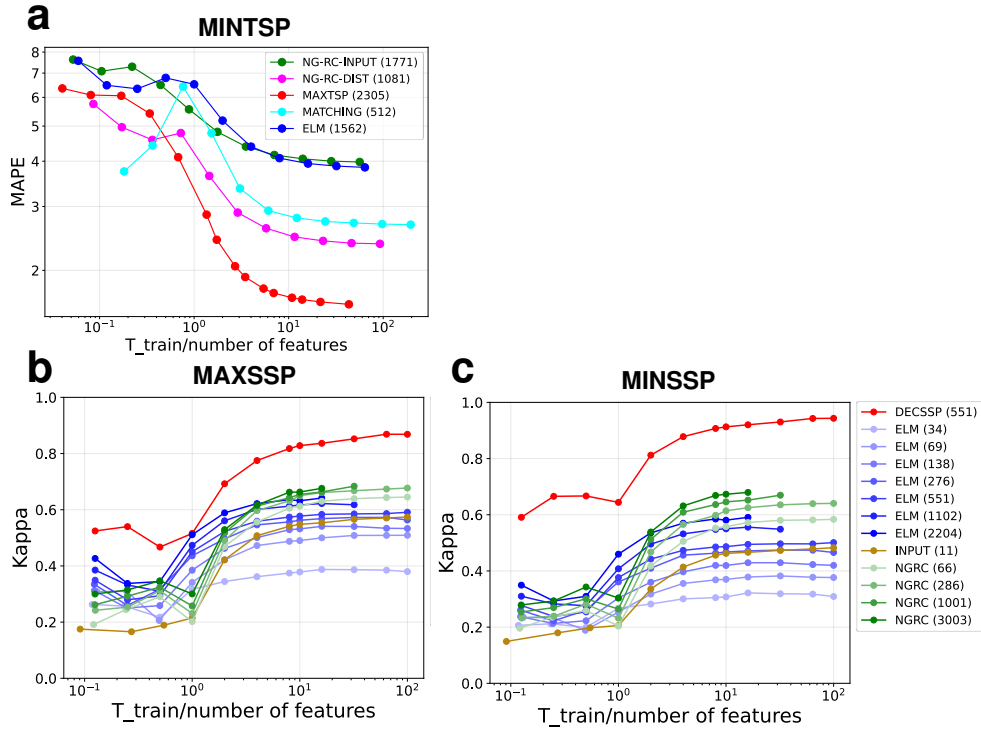}
\caption{\textbf{Training-size dependence of the optimal value prediction performance.} \textbf{a}, MAPE of the predicted MINTSP optimal value for the proposed method (MAXTSP) and baseline features (ELM, NG-RC-INPUT, NG-RC-DIST, MATCHING), as a function of the number of training samples divided by the number of features; the number of features used for each method is given in parentheses in the legend. \textbf{b}, \textbf{c}, Kappa of the predicted optimal value for \textbf{b}, MAXSSP and \textbf{c}, MINSSP, for the proposed method (DECSSP) and baseline features (INPUT, ELM, NGRC), as a function of the number of training samples divided by the number of features.}
\label{fig:training_size}
\end{figure}

To examine the required number of training samples, we conducted experiments in which the prediction performance was evaluated while varying the number of training samples.
The basic settings were the same as those of the experiments for predicting the optimal value of MINTSP and the optimal values of MAXSSP and MINSSP.
In addition to the proposed method, we evaluated various baseline methods; for methods whose number of features could be changed, we also varied the number of features.
For each method and each feature count, we plotted the performance, as the number of training samples was varied.
The results are shown in Fig.~\ref{fig:training_size}.
The horizontal axis is the number of training samples divided by the number of features.
Based on these results, we judged that using at least $10$ times as many training samples as features would be sufficient throughout this study.

\newpage
\section{Feature selection for MINTSP solution construction}
\begin{figure}[htbp]
\centering
\includesvg[width=0.8\linewidth]{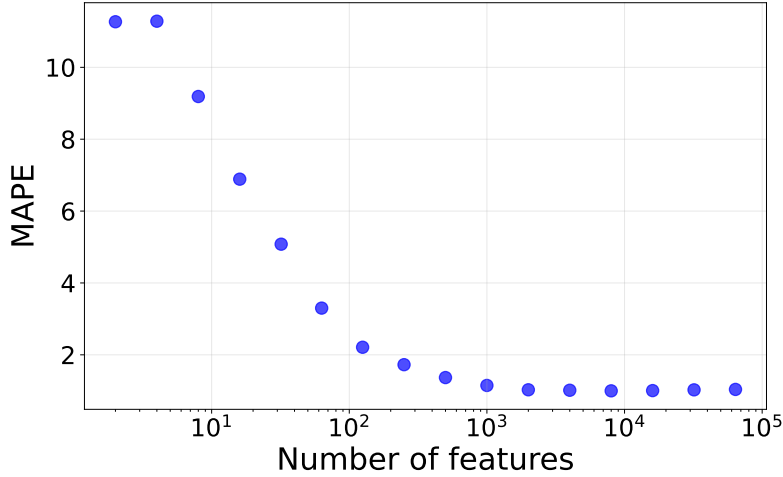}
\caption{\textbf{Effect of the number of selected MAXTSP-DP features on MINTSP solution construction performance.} The average percentage by which the tour constructed by the proposed method exceeds the optimal MINTSP tour length (MAPE) is shown as a function of the number of features selected from the MAXTSP DP table by recursive feature elimination.}
\label{fig:feature_selection_mintsp}
\end{figure}

In the MINTSP solution construction experiment, we varied the number of features selected from the MAXTSP DP table.
All other settings were kept the same.
The results, shown in Fig.~\ref{fig:feature_selection_mintsp}, reveal that that performance deteriorated as the number of features was reduced.
However, we found that the performance remained almost unchanged down to 2,000 features.
Because using fewer features shortens the computation time required to construct a solution, we selected 2,000 features in the main experiment, which was the smallest feature count among those that maintained performance.

\newpage
\section{Systematic analysis for the SSP experiments.}
We present detailed experimental results for SSP.

\begin{figure}[htbp]
\centering
\includesvg[width=1.0\linewidth]{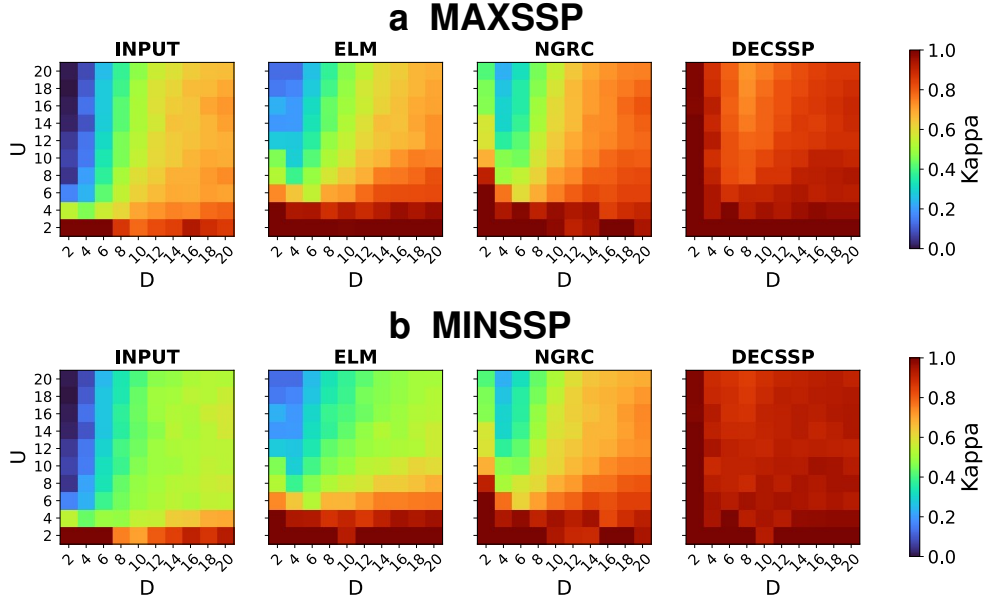}
\caption{\textbf{Optimal value prediction performance for SSP as the parameters $U$ and $D$ are varied.} \textbf{a}, \textbf{b}, Kappa of the predicted optimal value for \textbf{a}, MAXSSP and \textbf{b}, MINSSP, for the baseline features (INPUT, ELM, NGRC) and the proposed method (DECSSP), shown as a colormap as a function of the maximum input value $U$ and the input dimension $D$.}
\label{fig:ssp_change_ud}
\end{figure}

\paragraph{Optimal value prediction}
We examined how performance changed when the parameters $U$ and $D$ were varied.
We varied $U$ and $D$ from $2$ to $20$ in increments of $2$ and created colormaps using the normalized accuracy.
For each method, $T_{\mathrm{train}}$ was set to $10$ times its number of features, and $T_{\mathrm{eval}}=2,000$ was used throughout.
The number of features is $D$ for INPUT and $UD(D+1)/2$ for the proposed DECSSP method; for ELM, the number of features was matched to that of DECSSP, and for NG-RC, the polynomial degree was chosen so that its number of features was closest to that of DECSSP.
The results, shown in Fig.~\ref{fig:ssp_change_ud}, indicate that prediction becomes more difficult as $U$ increases.
However, the proposed method using DECSSP as features consistently achieved high performance.

\begin{figure}[htbp]
\centering
\includesvg[width=0.9\linewidth]{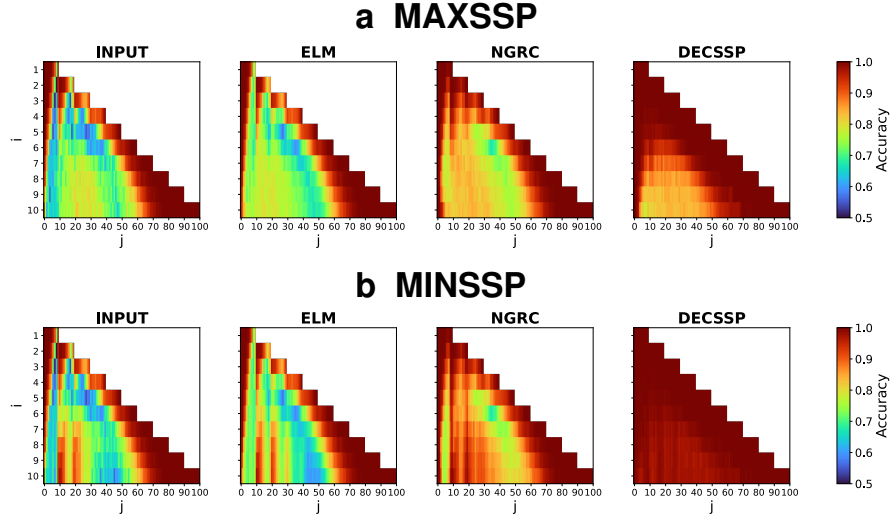}
\caption{\textbf{Accuracy of the objective functions predicted during SSP solution construction.} \textbf{a}, \textbf{b}, Accuracy of the predicted objective function $\mathrm{maxssp}_{i,j}(\mathbf{u})$ for \textbf{a}, MAXSSP and the corresponding objective function for \textbf{b}, MINSSP, for the baseline features (INPUT, ELM, NGRC) and the proposed method (DECSSP), shown as a colormap as a function of $i$ and $j$.}
\label{fig:ssp_construction_table}
\end{figure}

\begin{figure}[htbp]
\centering
\includesvg[width=0.85\linewidth]{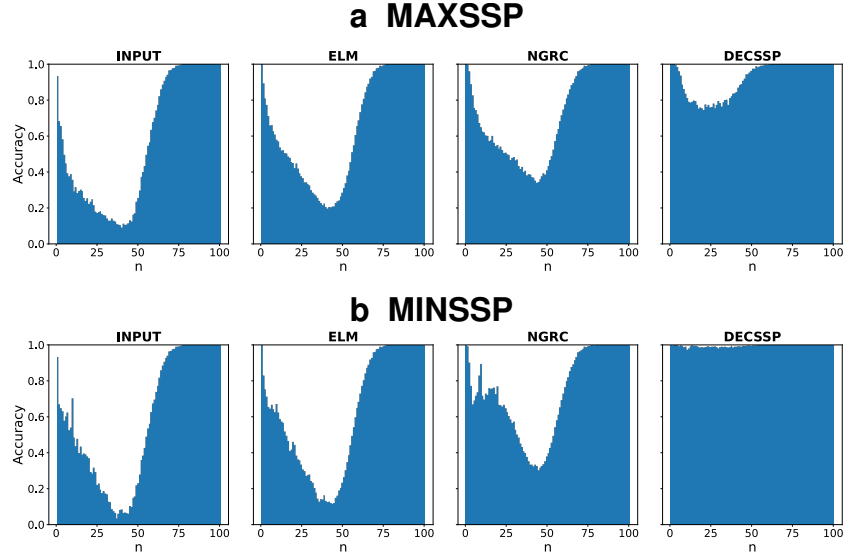}
\caption{\textbf{Solution construction accuracy for each target sum $n$ in SSP.} \textbf{a}, \textbf{b}, Accuracy of the constructed solutions for each target sum $n$, for \textbf{a}, MAXSSP and \textbf{b}, MINSSP, for the baseline features (INPUT, ELM, NGRC) and the proposed method (DECSSP).}
\label{fig:ssp_construction_bar}
\end{figure}

\paragraph{Solution construction}
We also compared the proposed method with the generic features used as baselines in the optimal value prediction experiments.
In addition, we examined how accurately the objective functions $\mathrm{maxssp}_{i,j}(\mathbf{u})$, predicted during solution construction, could be approximated.
For each objective function with $1 \leq i \leq D$ and $0 \leq j \leq UD$, the accuracy was visualized as a two-dimensional colormap.
The same analysis was conducted for MINSSP.
The results are shown in Fig.~\ref{fig:ssp_construction_table}.
The proposed method using DECSSP as features approximated the objective functions more accurately than the other methods.
We also confirmed that the prediction accuracy of the objective functions was consistent with the accuracy of solution construction.
Fig~\ref{fig:ssp_construction_bar} shows bar plots of the accuracy for each target sum $n$ in the solution construction experiment.
For large $n$, in most cases, the target sum $n$ exceeded the sum of all input elements, so no subset could be formed; this made the result trivial and yielded high accuracy for all methods.
Even in nontrivial cases, the proposed method constructed correct solutions with higher accuracy than the other methods.

\newpage
\bibliography{bib}

\end{document}